\documentclass[pdflatex,sn-basic,Numbered,super]{sn-jnl}

\usepackage{graphicx}%
\usepackage{multirow}%
\usepackage{amsmath,amssymb,amsfonts}%
\usepackage{amsthm}%
\usepackage{mathrsfs}%
\usepackage[title]{appendix}%
\usepackage{xcolor}%
\usepackage{textcomp}%
\usepackage{manyfoot}%
\usepackage{booktabs}%
\usepackage{algorithm}%
\usepackage{algorithmicx}%
\usepackage{algpseudocode}%
\usepackage{listings}%

\theoremstyle{thmstyleone}%
\theoremstyle{thmstyletwo}%

\theoremstyle{thmstylethree}%

\graphicspath{{figures/}}

\begin{document}

\title[Article Title]{Field-Space Autoencoder for Scalable Climate Emulators}


\author*[1]{\fnm{Johannes} \sur{Meuer}}\email{meuer@dkrz.de}

\author[1]{\fnm{Maximilian} \sur{Witte}}\email{witte@dkrz.de}
\equalcont{These authors contributed equally to this work.}

\author[1]{\fnm{Étiénne} \sur{Plésiat}}\email{plesiat@dkrz.de}
\equalcont{These authors contributed equally to this work.}

\author[1]{\fnm{Thomas} \sur{Ludwig}}\email{ludwig@dkrz.de}
\equalcont{These authors contributed equally to this work.}

\author[1]{\fnm{Christopher} \sur{Kadow}}\email{kadow@dkrz.de}
\equalcont{These authors contributed equally to this work.}

\affil*[1]{\orgdiv{Data Analysis}, \orgname{German Climate Computing Center (DKRZ)}, \orgaddress{\street{Bundesstr. 45a}, \city{Hamburg}, \postcode{20146}, \state{Hamburg}, \country{Germany}}}


\abstract{Kilometer-scale Earth system models are essential for capturing local climate change. However, these models are computationally expensive and produce petabyte-scale outputs, which limits their utility for applications such as probabilistic risk assessment. Here, we present the Field-Space Autoencoder, a scalable climate emulation framework based on a spherical compression model that overcomes these challenges. By utilizing Field-Space Attention, the model efficiently operates on native climate model output and therefore avoids geometric distortions caused by forcing spherical data onto Euclidean grids. This approach preserves physical structures significantly better than convolutional baselines. By producing a structured compressed field, it serves as a good baseline for downstream generative emulation. In addition, the model can perform zero-shot super-resolution that maps low-resolution large ensembles and scarce high-resolution data into a shared representation. We train a generative diffusion model on these compressed fields. The model can simultaneously learn internal variability from abundant low-resolution data and fine-scale physics from sparse high-resolution data. Our work bridges the gap between the high volume of low-resolution ensemble statistics and the scarcity of high-resolution physical detail.
}




\maketitle

\section{Introduction}
Climate modeling is currently aiming toward kilometer-scale resolution, which is necessary to capture small-scale physical processes such as convection, turbulence, and mesoscale dynamics \cite{Iles2019The,satoh2019global}. These high-resolution simulations are necessary for accurately predicting extreme weather events and local climate variability. However, managing the huge amounts of data they produce remains a challenge. A single simulation can generate petabytes of high-dimensional, multi-variable output \cite{Mizielinski2014Highresolution,segura2025nextgems,WarmWorldJSC,hoffmann2023destination}. These storage limitations and I/O bottlenecks limit the ability to easily share, analyze, and fully utilize these datasets \cite{Eggleton2020Open,Govett2024Exascale}. As the amount of high-resolution data increases, effective compression techniques are critical for balancing these constraints while preserving scientific fidelity.

Another motivation for data compression is the computational cost of generating the data in the first place. High-resolution simulations are computationally expensive, which limits the simulated time range or the ensemble size for uncertainty quantification \cite{Panta2025Scalable,Acosta2024The}. Consequently, the field is increasingly using machine learning–based emulation to generate ensembles for probabilistic risk assessment \cite{cachay2023dyffusion,price2023gencast}. However, training emulators on high-resolution data to learn fine-scale physics remains computationally intensive, and there is limited training data available. Therefore, compression alleviates storage constraints while enabling high-resolution physical information to be encoded into compact representations that support efficient training of generative models, such as latent diffusion models \cite{rombach2022high,meuer2024latent,zhuang2025ladcast,meuer2025latentensemble}. 

Convolutional architectures have been widely applied in climate science due to their broad applicability across diverse tasks and their strong spatial inductive biases, which enable the efficient learning of multiscale, locality-preserving features \cite{kadow2020artificial,meuer2024infilling,plesiat2024artificial}. Convolutional autoencoders (AEs) in particular have become a practical, data-driven approach for climate data compression, learning compact latent representations of high-dimensional fields that can be decoded with minimal information loss \cite{Guinness2016Compression,Fuglstad2020Compression,sha2020deep,Weylandt2023Beyond,Khan2024HighFidelity}. AEs can thus achieve reductions in data volume by orders of magnitude compared with classical lossless compression methods.

The convolutional models were adapted from image-processing models that use regular grids. However, modern climate models produce output data on spherical meshes, such as icosahedral grids, which better represent Earth's geometry. Using standard convolutions to project spherical data onto latitude–longitude grids introduces polar distortions, which degrade the reconstruction. Recent compression architectures, such as VAEformer \cite{han2024cra5} and hybrid Vision Transformers (ViTs) \cite{marzban2025hilab}, have shifted to attention-based techniques that address long-range dependencies more efficiently than standard convolutions. Furthermore, convolution-based models are inherently restricted to the spatial resolution on which they were trained and often rely on expanding feature spaces to capture complexity, which, in turn, increases computational overhead \cite{witte2025fieldspaceattentionstructurepreservingearth}. There is currently a lack of architecture that is: 1) consistent with Earth’s geometry, 2) resolution-invariant, and 3) computationally efficient enough to serve as a general-purpose standard for climate data compression and emulation. It is important to address these architectural limitations to bridge the gap in current climate data: extensive archives of low-resolution ensembles already exist and capture large-scale uncertainty, whereas high-resolution simulations remain scarce but are crucial for resolving fine-scale physical processes.

In this work, we aim to tackle these challenges through a framework that offers three primary contributions:

(1) We introduce the \emph{Field-Space Autoencoder} (FS-AE), a transformer-based architecture that achieves efficient climate data compression consistent with the Earth's spherical geometry. The model operates on the HEALPix grid \cite{gorski2005healpix} using Field-Space Attention \cite{witte2025fieldspaceattentionstructurepreservingearth} and Field-Space Compression/Decompression. We show that our model produces compressed representations that remain topologically faithful to the globe and outperforms state-of-the-art convolutional models.

(2) We demonstrate that this architecture is capable of \emph{Zero-Shot Super-Resolution}, enabling the processing of data at different scales within a single model. The multi-scale decomposition introduced by Witte et al. \cite{witte2025fieldspaceattentionstructurepreservingearth} decouples the architecture from a fixed grid size. This allows the model to generalize to resolutions on which it was not explicitly trained.

(3) We introduce \emph{Compressed Field Diffusion}, a generative model that applies a diffusion-based training on the compressed representations produced by the Field-Space Autoencoder. The resolution-invariance of the model enables a hybrid training strategy: we use large ensembles of low-resolution simulations to learn the large-scale internal variability of the climate while simultaneously learning from high-resolution data to capture fine-scale physical features. This alternative to standard latent diffusion demonstrates that ensembles of high-resolution climate simulations can be synthesized by combining the statistics of large, low-resolution ensembles with the structural detail of limited high-resolution data.

\begin{figure}[!ht]
\def\svgwidth{0.8\columnwidth}
\includegraphics[width=1.0\textwidth]{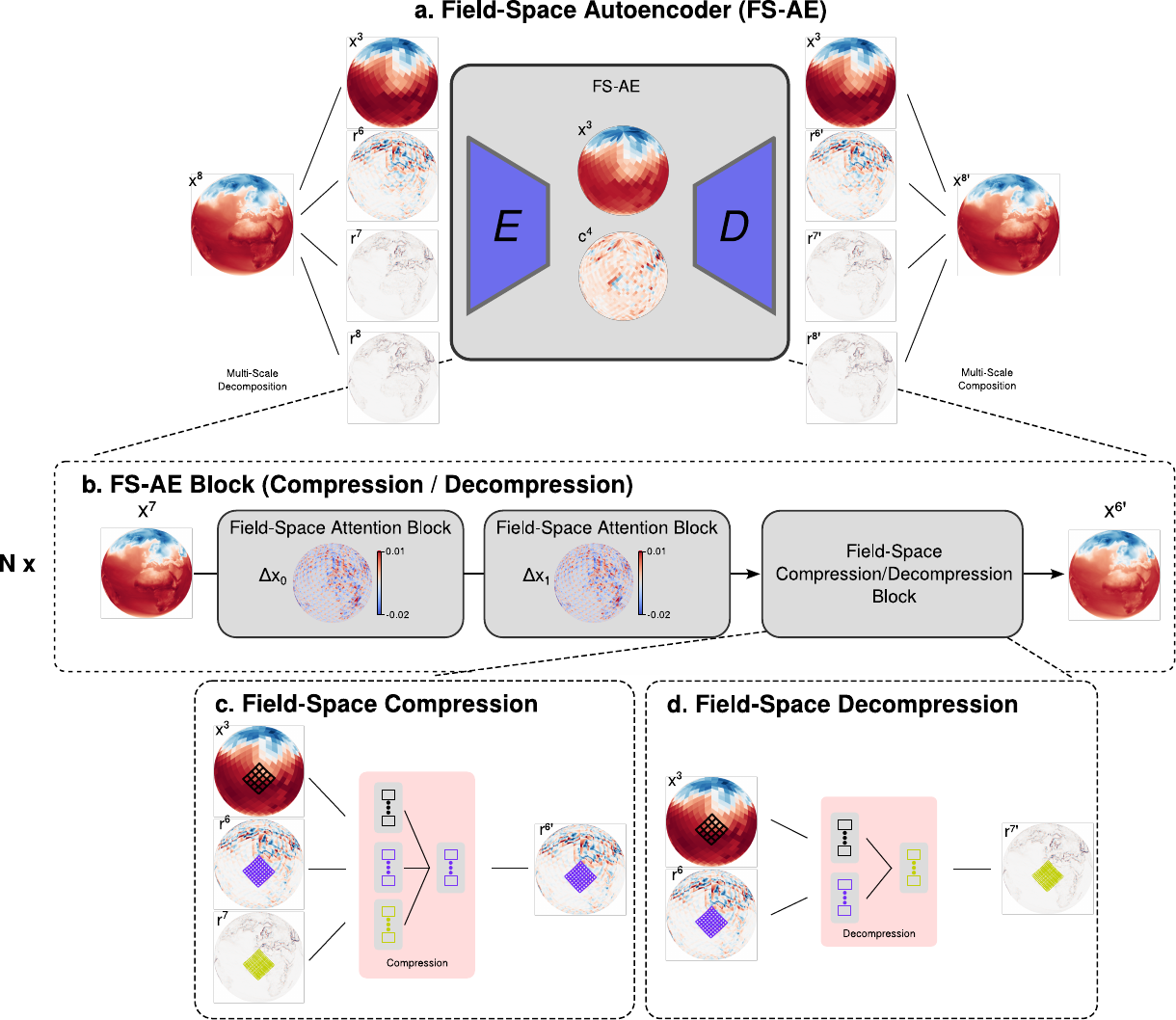}
\caption{\textbf{Field-Space Autoencoder: overview and multi-scale processing.}
(a) The Encoder $E$ compresses multi-scale inputs on the HEALPix sphere at a target coarse HEALPix level. The Decoder $D$ reconstructs fields at the original input HEALPix level. We denote native inputs by $x^{(n)}$, residuals by $r^{(n)}$, and compressed fields by $c^{(n)}$, where $n$ is the HEALPix level (resolution).
(b) Schematic Field-Space Autoencoder block combining Field-Space Attention and Field-Space Compression/Decompression (details in methods).
(c–d) Examples of a single Field-Space Compression/Decompression block: coarse base grid $x^3$ and residuals $r^6$/$r^7$ are aggregated and forwarded through a linear layer to produce the output at the requested residual level $r^{6'}$/$r^{7'}$.}
\label{fig:setup}
\end{figure}

\section{Results}

\bmhead{Field-Space Autoencoder enables efficient Large-Scale Data Compression}
To address the geometric distortions caused by projecting global data onto Euclidean grids, we developed a compression architecture that operates natively on the HEALPix spherical mesh \cite{gorski2005healpix}. Latent-space models, such as Vision Transformers (ViTs) and Convolutional Neural Networks (CNNs), learn to compress abstract feature vectors by embedding spatial patches into a latent representation. In our Field-Space Autoencoder, we leverage the field-space concept of the Field-Space Attention \cite{witte2025fieldspaceattentionstructurepreservingearth} to explicitly define compression/decompression in space (see Fig. \ref{fig:setup}c and d, see methods for details). By avoiding feature space expansion, the model achieves high computational efficiency and outputs a compressed field state with a feature dimension of one, rather than a multi-channel latent tensor.

Our Field-Space Autoencoder employs a multi-resolution residual strategy using multi-scale decomposition \cite{witte2025fieldspaceattentionstructurepreservingearth}: the input field is decomposed into a hierarchy of resolutions (see Methods). The model preserves the coarsest resolution as a base state, while higher resolutions are encoded as residuals relative to the coarser levels. This hierarchical approach induces resolution invariance, allowing the same model to process inputs of varying granularity and enabling zero-shot super-resolution by simply masking fine-scale residuals during inference.

\begin{figure}[!ht]
\def\svgwidth{0.72\linewidth}
\centering
\footnotesize{
\begingroup%
  \makeatletter%
  \providecommand\color[2][]{%
    \errmessage{(Inkscape) Color is used for the text in Inkscape, but the package 'color.sty' is not loaded}%
    \renewcommand\color[2][]{}%
  }%
  \providecommand\transparent[1]{%
    \errmessage{(Inkscape) Transparency is used (non-zero) for the text in Inkscape, but the package 'transparent.sty' is not loaded}%
    \renewcommand\transparent[1]{}%
  }%
  \providecommand\rotatebox[2]{#2}%
  \newcommand*\fsize{\dimexpr\f@size pt\relax}%
  \newcommand*\lineheight[1]{\fontsize{\fsize}{#1\fsize}\selectfont}%
  \ifx\svgwidth\undefined%
    \setlength{\unitlength}{190.6312294bp}%
    \ifx\svgscale\undefined%
      \relax%
    \else%
      \setlength{\unitlength}{\unitlength * \real{\svgscale}}%
    \fi%
  \else%
    \setlength{\unitlength}{\svgwidth}%
  \fi%
  \global\let\svgwidth\undefined%
  \global\let\svgscale\undefined%
  \makeatother%
  \begin{picture}(1,1.62706981)%
    \lineheight{1}%
    \setlength\tabcolsep{0pt}%
    \put(0,0){\includegraphics[width=\unitlength,page=1]{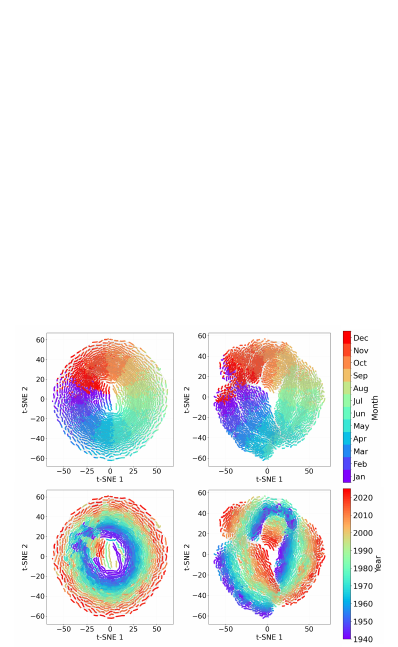}}%
    \put(0.49407019,1.58023474){\color[rgb]{0,0,0}\makebox(0,0)[t]{\lineheight{1.25}\smash{\begin{tabular}[t]{c}\textbf{\scriptsize{a. Compression Ratio vs weighted RMSE and PSNR}}\end{tabular}}}}%
    \put(0.49026369,0.84429751){\color[rgb]{0,0,0}\makebox(0,0)[t]{\lineheight{1.25}\smash{\begin{tabular}[t]{c}\textbf{\scriptsize{b. t-SNE Analysis on Compressed / Latent Variables}}\end{tabular}}}}%
    \put(0.27432989,0.79265745){\color[rgb]{0,0,0}\makebox(0,0)[t]{\lineheight{1.25}\smash{\begin{tabular}[t]{c}FS-AE\end{tabular}}}}%
    \put(0.08364537,0.40528257){\color[rgb]{0,0,0}\makebox(0,0)[t]{\lineheight{1.25}\smash{\begin{tabular}[t]{c}Outliers\end{tabular}}}}%
    \put(0.68434796,0.79162027){\color[rgb]{0,0,0}\makebox(0,0)[t]{\lineheight{1.25}\smash{\begin{tabular}[t]{c}CNN-VAE\end{tabular}}}}%
    \put(0,0){\includegraphics[width=\unitlength,page=2]{eval_tas.pdf}}%
  \end{picture}%
\endgroup%
}
\caption{\textbf{Reconstruction performance and compressed space visualization.} (a) Dual-axis plot showing root-mean-square error (RMSE; left vertical axis, °C) and peak signal-to-noise ratio (PSNR; right vertical axis, dB) as functions of compression ratio for both the Field-Space Autoencoder (FS-AE) and CNN-VAE models. The effective compression ratios of our models are slightly lower compared to the corresponding CNN-VAE models, because we additionally maintain an average field at the coarsest resolution (see Fig. \ref{fig:setup}). (b) Two-dimensional t-SNE projection of the compressed/latent representations (for $f=16$) of the full daily ERA5 near-surface air temperature (tas) dataset covering 1940–2024, computed with a fixed random seed for reproducibility. The same projection is shown in two views: colored by month (top) and by year (bottom). Each point corresponds to one encoded daily field.}
\label{fig:tas_eval}
\end{figure}

We first evaluate this architecture on surface air temperature ($tas$) from ERA5\cite{hersbach2020era5}, a global reanalysis dataset at 0.25° ($\approx$ 30km) resolution, which is widely considered the gold standard for evaluating ML models for climate prediction\cite{lam2023graphcast,bi2023panguweather,kurth2023fourcastnet,price2023gencast}. We perform this evaluation across four compression ratios ($f \in \{16, 64, 256, 1024\}$). As a baseline, we employ a state-of-the-art Convolutional VAE (CNN-VAE) with global spatial attention\cite{rombach2022high}, which represents the current standard for learned climate data compression. Our transformer consistently outperforms the CNN-VAE baseline in reconstruction fidelity across all compression ratios (Fig. \ref{fig:tas_eval}a). The Field-Space Autoencoder's reconstruction error at $64\times$ compression (RMSE $\approx$ 0.3 °C) is lower than that of the CNN-VAE at only $16\times$ compression, meaning it achieves 4x higher compression efficiency while maintaining the same accuracy across all metrics. Even under extreme compression ($f=256$ and $f=1024$), we retain a high PSNR, whereas the baseline degrades significantly.

\bmhead{Compressed Space Encodes Physical Dynamics}
A well-organized latent space is essential for downstream capabilities, such as outlier detection and generative sampling \cite{Connor2020Variational,Angiulli2022}. To examine this property, we analyze the learned compressed space using a 2D t-SNE projection \cite{arora2018analysis} of the encoded daily fields from 1940–2024 (Fig. \ref{fig:tas_eval}b).

The Field-Space Autoencoder embedding reveals an organized compressed field by highlighting physically meaningful relationships rather than just pixel statistics. The top row of Fig. \ref{fig:tas_eval}b shows that the compressed states arrange themselves by season, forming a continuous, cyclic trajectory that mirrors the Earth's annual cycle. When colored by year (bottom row), the analysis shows a radial progression that extends from the cooler mid-20th century at the center to the warmer present-day at the periphery. This indicates that our model has implicitly learned to preserve the signal of global warming without supervision.

Furthermore, the embedding structure offers a plausible physical interpretation for specific anomalies. The outliers (highlighted in black) located deep within the central region of the projection coincide with major volcanic eruptions such as El Chichón (1982) and Mount Pinatubo (1991) around that time. Given that cooler years are concentrated in the center, the clustering of these events is consistent with the global cooling that they induced.

In contrast, the CNN-VAE latent space does not exhibit the clean, organized structures in the t-SNE projection, even though it is explicitly regularized by a KL loss term. This demonstrates that the Field-Space Autoencoder naturally yields a well-organized representation, whereas the baseline struggles to capture such physical dependencies.

\begin{figure}[!ht]
\def\svgwidth{\linewidth}
\centering
\footnotesize{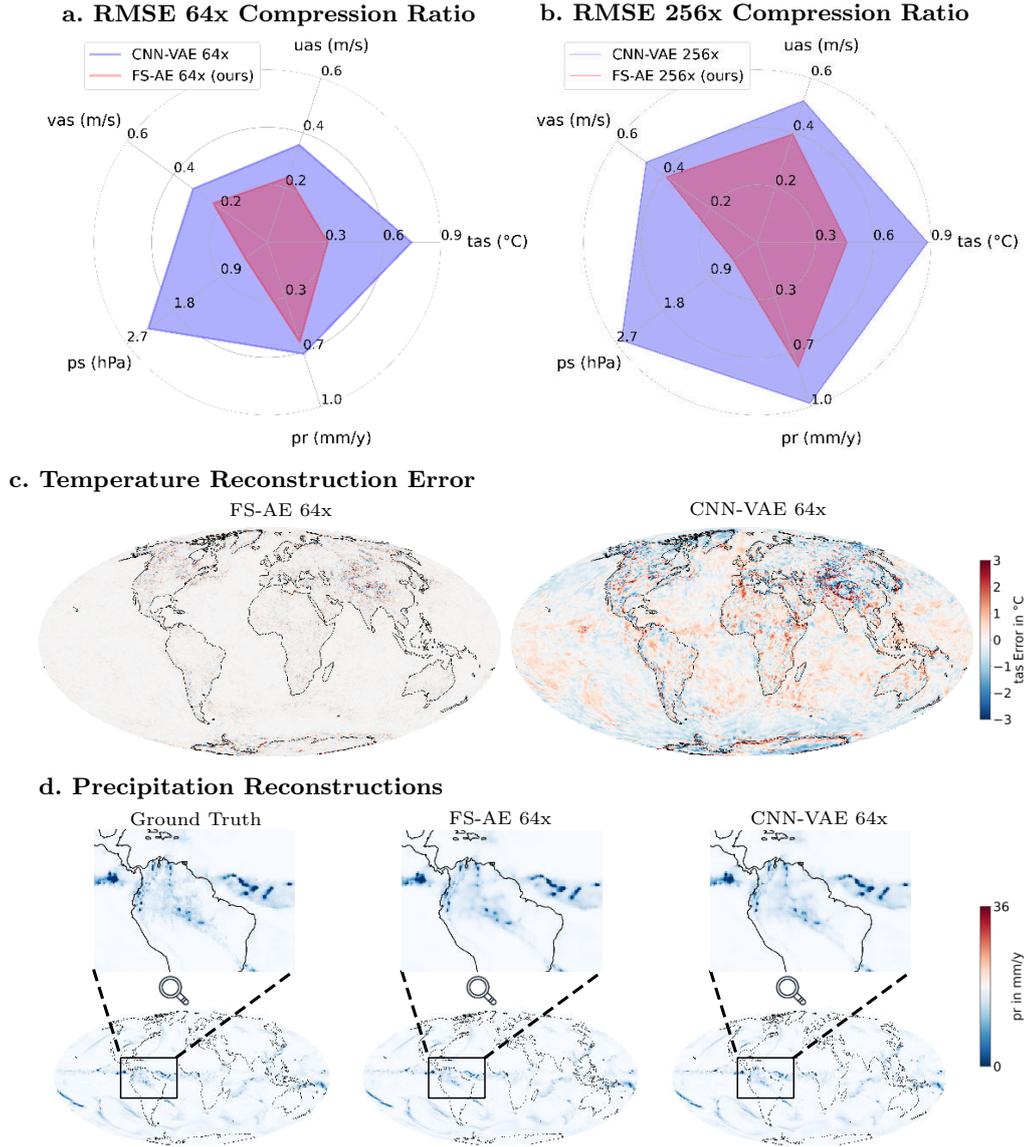}
\caption{\textbf{Multi-variable reconstruction evaluation.} (a-b) Radar plot showing per-variable RMSE at respective compression for five ERA5 variables: surface air temperature (tas), eastward and northward wind at 10m (uas, vas), surface pressure (ps), and precipitation (pr). Values closer to the center of the radar chart correspond to lower RMSE. (c) Global reconstruction error of a selected sample at 64× compression. Positive values indicate too warm predictions, negative too cold. (d) Ground Truth and reconstructions of global precipitation of a selected sample at 64× compression.}
\label{fig:multivar_eval}
\end{figure}

\bmhead{Scalable Multi-Variable Adaptation}
An advantage of our transformer architecture is its modularity. While CNN-based architectures require retraining from scratch for new variable combinations, our transformer allows for efficient fine-tuning. We extend the pre-trained single-variable models, each with its own set of learnable parameters, with cross-variable attention layers at the bottleneck (similar to Lessig et al. \cite{lessig2023atmorep}) to process five fields simultaneously: temperature ($tas$), wind components ($uas, vas$), surface pressure ($ps$), and precipitation ($pr$). Figure \ref{fig:multivar_eval}a-b) shows the per-variable RMSE performance over the retrained CNN-VAE baseline at compression ratios $f=64$ and $f=256$, respectively. The reconstruction error is around 50\% for temperature and 30\% for winds, compared to the CNN-VAE. We confirm this quantitative advantage visually by the temperature error maps (Fig. \ref{fig:multivar_eval}c). Our Field-Space Autoencoder limits errors to regions with high gradients, such as mountains, where the CNN-VAE produces even higher errors. Additionally, the CNN-VAE struggles to reconstruct smoother patterns, such as those in the ENSO region. Our model, however, performs quite well in these regions. Precipitation (Fig. \ref{fig:multivar_eval}d) remains the most challenging variable for both architectures. Due to its stochastic, highly non-linear nature, deterministic compression yields smoothed representations, as confirmed by other studies\cite{shiraishi2025wasserstein}. However, unlike the baseline, which appears to over-prioritize precipitation at the expense of other variables, our model maintains a balanced fidelity across all variables.

\bmhead{Zero-Shot Super-Resolution}
The multi-resolution residual design of our Field-Space Autoencoder, via multi-scale decomposition, decouples the architecture from a fixed grid size. This allows the model to map inputs of different scales into a shared compressed space and decode them onto the high-resolution grids it was trained on without fine-tuning. We evaluate this zero-shot capability in two contexts: super-resolution of coarsened ERA5 data and of lower-resolution climate model output.

\begin{figure}[!ht]
\def\svgwidth{\linewidth}
\centering
\footnotesize{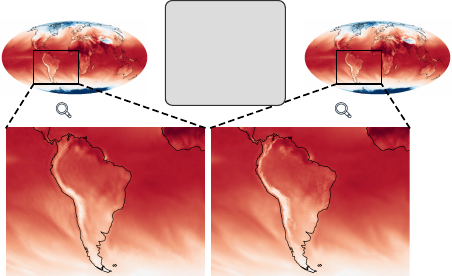}
\caption{\textbf{Compression and Zero-shot super-resolution on the MPI-ESM historical simulations.} Left: Original snapshots of surface temperature ($tas$) from the MPI-ESM historical simulation (HEALPix level 6). Right: The corresponding output from the Field-Space Autoencoder ($64\times$ compression) after compressing the input and decoding to level HEALPix 8.}
\label{fig:sr_eval}
\end{figure}

To evaluate zero-shot super-resolution, we mask the fine-scale residual tokens during inference. We simulate a $4\times$ upscaling task by masking level 8 residuals ($r^8$; input Level 7 $\to$ output Level 8) and a $16\times$ upscaling task by masking both level 7 and 8 residuals ($r^7, r^8$; input Level 6 $\to$ output Level 8). The reconstruction error increases with the upscaling factor, but the model still maintains reasonable accuracy (see Appendix Fig. \ref{fig:radar_sr}) and high-frequency features (see Appendix Fig. \ref{fig:spetral_analysis_sr}). This confirms that the model synthesizes missing fine-scale details by using the physical priors learned during high-resolution training, rather than simply interpolating the coarse input.

This capability extends beyond simple reconstruction to the downscaling of climate simulations. We apply our pretrained model ($f=64$) to historical simulations from the MPI-ESM1.2-HR model (Figure \ref{fig:sr_eval}). The original data, which is natively given at $\approx 100$ km resolution (mapped via distance weighting to HEALPix level 6), serves as input, while the finer residual levels ($r^7$ and $r^8$) are set to zero.

As with the downsampled ERA5 data, after compressing the field, the model successfully synthesizes high-frequency structures that were absent in the source data. We show this effect for surface temperature ($tas$): the model output (right) exhibits sharper gradients and more topographic detail than the smoother original simulation (left). This can also be observed in other variables (see Appendix Fig. \ref{fig:eval_sr_allvar}). The spectral analyses of surface temperatures (see Appendix Fig. \ref{fig:spetral_analysis_temp_mpiesm}) and zonal winds (see Appendix Fig. \ref{fig:spetral_analysis_uas_mpiesm}) confirm this trend. The MPI-ESM model shows a falloff in higher frequencies, whereas the Field-Space Autoencoder maintains them with only a slight decrease relative to ERA5, on which it was trained. This demonstrates that our model not only acts as a compression tool but also as a bridge between resolutions. It effectively upgrades coarser climate model outputs to a pseudo-high-resolution state while simultaneously reducing data volume.

\bmhead{Generative Emulation via Compressed Field Diffusion}
Finally, we demonstrate that the compressed field serves as a robust foundation for downstream climate emulation. We introduce \emph{Compressed Field Diffusion}, a generative diffusion framework that operates directly on the compressed field.

In this final experiment, we employ the pre-trained multi-variable Field-Space Autoencoder with a compression ratio of $f=64$. To prepare the training data for the diffusion model, we project five atmospheric variables (\emph{tas}, \emph{uas}, \emph{vas}, \emph{ps}, \emph{pr}) from a 10-member ensemble of MPI-ESM1.2-HR historical simulations ($\approx 100$ km resolution) into the compressed field space via the Encoder. We then train the diffusion model to generate sequences within this compressed domain, conditioning on timestamps to produce temporally consistent trajectories over selected time ranges. The diffusion and denoising processes are defined separately for the base-level 3 average and the learned compressed residuals (see Methods). For evaluation, we synthesize a new 10-member ensemble in the compressed field, decode these fields to the native ERA5 resolution ($\approx 25$ km) using the Field-Space Autoencoder decoder, and compare the results with the original MPI-ESM1.2-HR ensemble.

\begin{figure}[!ht]
\def\svgwidth{\linewidth}
\centering
\footnotesize{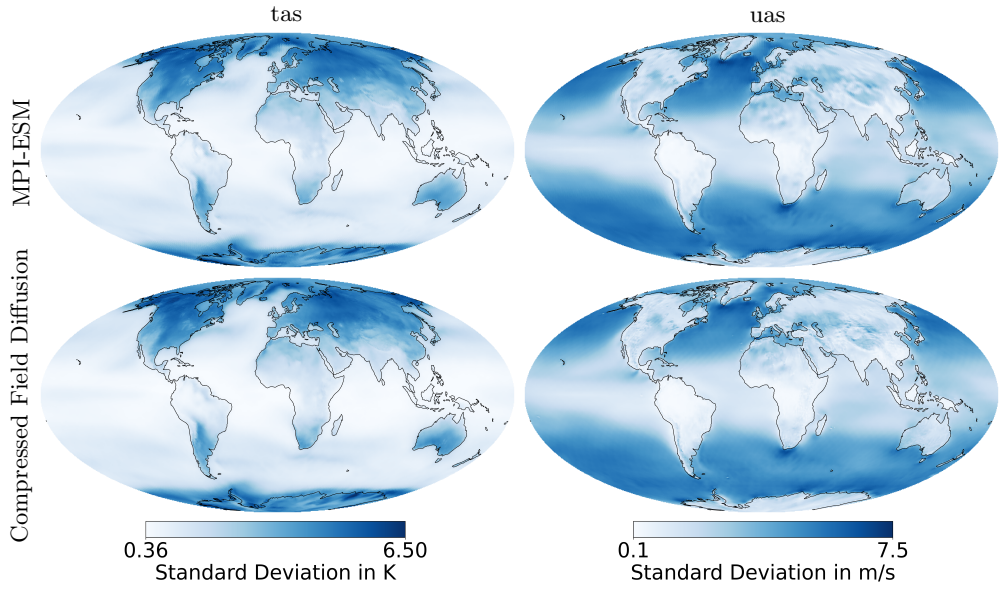}
\caption{\textbf{Evaluation of generative fidelity and ensemble variability.} a. Global spherical power spectral density of a subset of the output variables ($tas$ and $uas$) of a single simulation over a year. b. Maps of internal variability (standard deviation) for the original 10-member MPI-ESM1.2-HR ensemble (left) and the synthetic 10-member ensemble generated via Compressed Field Diffusion (right). The standard deviation was calculated over the 10 ensemble members and then averaged over a full year.}
\label{fig:diffusion_eval}
\end{figure}

Spectral analysis (Fig. \ref{fig:diffusion_eval}a) reveals that the diffusion-generated fields reconstruct high frequencies absent in the MPI-ESM output. Although the generated spectrum does not exactly match the spectrum of the ERA5 data, it more closely aligns with the reanalysis than with the low-resolution climate model output. At lower frequencies, there is a slight underestimation compared to the original MPI-ESM ensemble. This suggests that the diffusion model could be improved, as this underestimation could not be observed when the Field-Space Autoencoder was given the MPI-ESM data as input (see Fig. \ref{fig:spetral_analysis_temp_mpiesm} and \ref{fig:spetral_analysis_uas_mpiesm}). Nevertheless, the generated ensemble preserves the statistics of the original historical ensemble. As shown in Fig. \ref{fig:diffusion_eval}b, the internal variability of our synthetic ensemble replicates the spatial variance patterns of the source MPI-ESM ensemble. This confirms that \emph{Compressed Field Diffusion}, combined with our resolution-invariant autoencoder, can increase the information content of legacy ensembles by synthesizing simulations that are spectrally sharp while remaining statistically consistent with the source model's climate variability.

\section{Conclusion}

In this work, we redefined climate data compression and emulation with the Field-Space Autoencoder. The proposed framework achieved the three primary contributions:

(1) By leveraging Field-Space Attention and Compression on the HEALPix mesh, our Field-Space Autoencoder avoids the computationally expensive feature-space expansion typical of standard deep learning models and eliminates projection artifacts. 

(2) We provided a resolution-invariant framework. Our multi-scale design preserves the physical structure of the atmosphere and enables zero-shot super-resolution. This allows the model to generalize across different data scales.

(3) We demonstrated that the Field-Space Autoencoder provides a robust foundation for emulation. Our \emph{Compressed Field Diffusion} framework leverages this lightweight compressed space to bridge the scale gap. This enables a hybrid training strategy that combines the internal variability of large, low-resolution ensembles with the spectral sharpness of high-resolution data.

Future iterations must extend beyond the five variables tested here in order to account for further correlations between variables and pressure levels. A critical next step is to address the smoothing of highly stochastic fields, such as precipitation, by replacing deterministic with probabilistic objectives that can capture fine-scale climate phenomena. Additionally, the temporal redundancy in climate data could enable even higher compression ratios when compressing in time.

Our Field-Space Autoencoder provides a blueprint for creating a unified representation of climate data that can be stored, shared, and used to train the next generation of physics-aware foundation models efficiently.

\backmatter


\bmhead{Acknowledgments}
The authors are grateful to the German Climate Computing Center (DKRZ) for providing the hardware for the calculations and all the data that were used for this study. Johannes Meuer is funded from the German National funding agency (DFG), provided by the research unit FOR 2820, titled "Revisiting The Volcanic Impact on Atmosphere and Climate-Preparations for the Next Big Volcanic Eruption" (VolImpact), with the project number 398006378. This project received funding from the German Federal Ministry of Research, Technology and Space (BMBFTR) under grant 16ME0679K. Supported by the European Union–NextGenerationEU; and in part by the Horizon Europe project EXPECT under Grant No. 101137656.

\clearpage

\graphicspath{{figures/}}

\section{Methods}

\subsection{Data}\label{subsec:data}

\bmhead{Grids and Resolution}

We use two spherical discretizations: (i) the native ERA5 regular latitude–longitude grid (\(\sim0.25^{\circ}\) spacing), and (ii) an equal‑area HEALPix grid for token‑based models. We index HEALPix resolution by a discrete zoom level \(z\in\mathbb{Z}_{\ge0}\) with
\[
N_{\text{side}}(z)=2^{z}, 
\qquad
N_{\text{pix}}(z)=12\cdot 4^{z},
\]
so that each pixel at level \(z\) subdivides into four children at \(z{+}1\). Unless otherwise stated, we use \(z{=}8\) (\(N_{\text{side}}{=}256\), \(N_{\text{pix}}{=}786{,}432\)), which provides an effective angular resolution comparable to the ERA5 \(0.25^{\circ}\) grid while guaranteeing equal‑area cells and iso‑latitude rings \cite{gorski2005healpix}. We denote the coarsest and finest HEALPix levels used in the model by \(z_{\min}\) and \(z_{\text{in}}\).

\bmhead{ERA5 Dataset}

We use ECMWF ERA5 reanalysis at hourly cadence and aggregate each variable to daily means before any normalization or train/test splits. Convolutional baselines (CNN–VAE family) are trained/evaluated on the native \(0.25^{\circ}\) latitude–longitude grid. The Field-Space Autoencoder models are trained/evaluated on ERA5 remapped to the HEALPix sphere at \(z{=}8\).

For the lat‑lon \(\rightarrow\) HEALPix remapping, each HEALPix pixel center receives an inverse‑distance weighted average of nearby regular‑grid samples computed using great‑circle distances; weights are normalized to sum to one. This preserves large‑scale means while avoiding latitude‑dependent area bias on the source grid.

All compression models are trained on 1940–2021 and evaluated on a temporal hold‑out from 2022–2025 (through April 2025). No evaluation‑year samples are used during training, and all preprocessing statistics (below) are computed exclusively on the training period.

\bmhead{MPI-ESM1.2-HR Historical Ensemble}
To enable hybrid training and evaluate the model's ability to bridge climate model resolutions, we utilize data from the Max Planck Institute Earth System Model (MPI-ESM1.2) in its High-Resolution (HR) configuration\cite{Müller2018A}. We select a 10-member ensemble of historical simulations covering the period 1940–2014, which follow the CMIP6 historical forcing protocols. The atmospheric component (ECHAM6.3) of the HR configuration operates at a horizontal resolution of T127 (approximately 100,km or 0.93$^{\circ}$ at the equator) with 95 vertical levels. For consistency with our geometric architecture, we regrid the native spectral output to the HEALPix mesh at level 6 ($N_{\text{side}}=64$), following the same remapping process as for the ERA5 dataset. The level 6 resolution corresponds to an approximate resolution of 100,km, which serves as the coarse-grain input for our downscaling and hybrid diffusion experiments. The \emph{Compressed Field Diffusion} results were produced using a one-year prediction and were compared with the corresponding MPI-ESM1.2-HR simulations from the year 1940.

\bmhead{Preprocessing and Normalization}

Daily aggregation is followed by a per‑variable, percentile‑based scaling using only training data. Let \(p01_v\) and \(p99_v\) denote the 1st and 99th percentiles of variable \(v\) on the training split. We scale
\[
\tilde{x}^{(v)}=\frac{x^{(v)}-p01_v}{\,p99_v-p01_v\,},
\]
without clipping; values outside \([p01_v,p99_v]\) may map outside \([0,1]\). After inference/decoding, we invert the same transform to recover physical units. All error metrics are reported in physical units.

\bmhead{Area‑Consistent Evaluation}

Because cell areas vary with latitude on the regular grid, scalar error metrics there use latitude weights \(w_{ij}=\cos\varphi_i\) (with \(\varphi_i\) the latitude of row \(i\)) following common practice \cite{han2024cra5,zhuang2025ladcast}. For example, the latitude‑weighted RMSE is
\[
\mathrm{RMSE}_{w}
=\left(
\frac{\sum_{t}\sum_{i,j} w_{ij}\,\big(x_{t,ij}-\hat{x}_{t,ij}\big)^{2}}
{\sum_{t}\sum_{i,j} w_{ij}}
\right)^{1/2}.
\]
On HEALPix (equal‑area) we use unweighted spatial means, which are area‑consistent by construction.

\subsection{Model Architecture}

We adopt the multi–scale decomposition, scale conservation, and field-space attention design introduced in the published work \emph{Field-space attention for structure-preserving earth system transformers}\cite{witte2025fieldspaceattentionstructurepreservingearth}. Our implementation mirrors that approach and we refer readers to  Witte et al.\cite{witte2025fieldspaceattentionstructurepreservingearth} for complete derivations, ablations, and implementation details. We introduce Field-Space Compression and Field-Space Decompression as complementary techniques to the Field-Space Attention by Witte et al. for compression and decompression, respectively.

\bmhead{Field-Space Compression and Decompression}

Each Compression and Decompression block performs a linear mapping \emph{per patch} across selected zoom levels, using a user‑defined \emph{patch zoom} \(z_P\) that sets the patch size. A patch anchored at \(z_P\) contains
\[
n_{\mathrm{pix}}(z\mid z_P)=
\begin{cases}
4^{(z-z_P)}, & z\ge z_P \quad\text{(children grouped into channels)},\\
1, & z< z_P \quad\ \ \text{(ancestor broadcast)}
\end{cases}
\]
pixels from level \(z\).

For generating the input patches, let \(\mathcal{Z}_{\mathrm{in}}\subseteq\{z_{\min},\dots,z_{\mathrm{in}}\}\) be the set of input levels (typically several residual levels \(r^{(z)}\) plus one coarse mean level \(x^{(z_c)}\)). For each token on the \(z_P\)-grid, we reshape each included level onto the patch and concatenate along the channel axis:
\[
P^{(z_P)} \;=\;
\operatorname{Concat}\Bigl(
\{\mathcal{R}_{z\to z_P}(r^{(z)}) : z\in \mathcal{Z}_{r}\}
\;\cup\;
\{\mathcal{R}_{z_c\to z_P}(x^{(z_c)})\}
\Bigr)
\in \mathbb{R}^{B\times N_{z_P}\times C_{\mathrm{in}}},
\]
with
\[
C_{\mathrm{in}}=\sum_{z\in\mathcal{Z}_{r}\cup\{z_c\}} n_{\mathrm{pix}}(z\mid z_P),
\qquad
\mathcal{R}_{z\to z_P}(\cdot)\in\mathbb{R}^{B\times N_{z_P}\times n_{\mathrm{pix}}(z\mid z_P)}.
\]
We keep one channel per physical field, \(C{=}1\); scale mixing occurs only within \(P^{(z_P)}\).

For compression/downsampling, let \(z_{\max}=\max \mathcal{Z}_{r}\) be the highest residual level provided to the block and set the target to the next coarser level \(z_{\downarrow}=z_{\max}-1\). A shared linear map is applied per patch:
\[
H^{\Downarrow} \;=\; W^{\Downarrow} P^{(z_P)} + b^{\Downarrow}
\ \in\ \mathbb{R}^{B\times N_{z_P}\times C_{\mathrm{out}}^{\Downarrow}},
\qquad
W^{\Downarrow}\in\mathbb{R}^{\,C_{\mathrm{out}}^{\Downarrow}\times C_{\mathrm{in}}},
\]
with the number of output channels equal to the number of pixels \emph{in the target level’s patch}:
\[
C_{\mathrm{out}}^{\Downarrow}=n_{\mathrm{pix}}(z_{\downarrow}\mid z_P)=4^{\max(0,\;z_{\downarrow}-z_P)}.
\]
A reshape distributes the patch outputs onto the \(z_{\downarrow}\)-grid:
\[
\widehat{r}^{(z_{\downarrow})}=\mathcal{H}_{z_P\to z_{\downarrow}}\!\bigl(H^{\Downarrow}\bigr)
\ \in\ \mathbb{R}^{B\times N_{z_{\downarrow}}\times 1},
\]
where \(\mathcal{H}_{z_P\to z}\) is the canonical HEALPix parent–child unstacking satisfying \(N_{z_P}\!\cdot n_{\mathrm{pix}}(z\mid z_P)=N_z\).
All levels strictly lower than \(z_{\downarrow}\) (e.g., \(z_c\)) are forwarded unchanged together with \(\widehat{r}^{(z_{\downarrow})}\); the highest level \(z_{\max}\) is removed at this step.

\emph{Example (See Fig. \ref{fig:setup}c):} If \(z_P{=}1\) and inputs are \((z{=}7,6)\) residuals plus \(x^{(3)}\), then
\[
C_{\mathrm{in}}=n_{\mathrm{pix}}(7\mid 1)+n_{\mathrm{pix}}(6\mid 1)+n_{\mathrm{pix}}(3\mid 1)=4096+1024+16=5136,
\]
and compressing \(7\to 6\) yields \(C_{\mathrm{out}}^{\Downarrow}=n_{\mathrm{pix}}(6\mid 1)=1024\).
This realizes the mapping \(\{7,6,3\}\to \{6\}\).

For decompression/upsampling, given inputs up to \(z_{\max}\), set \(z_{\uparrow}=z_{\max}+1\). The per‑patch linear map produces the target‑level patch, which is then scattered to the finer grid:
\[
H^{\Uparrow} \;=\; W^{\Uparrow} P^{(z_P)} + b^{\Uparrow}
\ \in\ \mathbb{R}^{B\times N_{z_P}\times C_{\mathrm{out}}^{\Uparrow}},
\quad
C_{\mathrm{out}}^{\Uparrow}=n_{\mathrm{pix}}(z_{\uparrow}\mid z_P),
\]
\[
\widehat{r}^{(z_{\uparrow})}=\mathcal{H}_{z_P\to z_{\uparrow}}\!\bigl(H^{\Uparrow}\bigr)
\ \in\ \mathbb{R}^{B\times N_{z_{\uparrow}}\times 1}.
\]
All levels at or below \(z_{\max}\) pass through unchanged and are concatenated with \(\widehat{r}^{(z_{\uparrow})}\) for the next layer.

\emph{Example (See Fig. \ref{fig:setup}d):} With \(z_P{=}1\) and inputs at \(\{z{=}6\}\) plus \(x^{(3)}\), we have
\[
C_{\mathrm{in}}=n_{\mathrm{pix}}(6\mid 1)+n_{\mathrm{pix}}(3\mid 1)=1024+16=1040,
\]
and upsampling \(6\to 7\) yields \(C_{\mathrm{out}}^{\Uparrow}=n_{\mathrm{pix}}(7\mid 1)=4096\).
This realizes the mapping \(\{6,3\}\to \{7\}\).

\subsection{Models}

\bmhead{Field-Space Autoencoder Models}
The Field-Space Autoencoder models take as input the mean field at a coarse HEALPix level together with three residual levels from the multi-scale decomposition described by Witte et al.\cite{witte2025fieldspaceattentionstructurepreservingearth}. For compression ratios \(16, 64,\) and \(256\), the mean field is provided at level \(z{=}3\), and for compression ratio \(1024\) at level \(z{=}1\). In all cases, residuals are taken at levels \(z{=}6,7,8\) on HEALPix. Each variant implements \(N\) Field-Space Autoencoder blocks in the encoder (using the two Field-Space Attention blocks followed by field-space compression block) and \(N\) Field-Space Autoencoder blocks in the decoder (using the two Field-Space Attention blocks followed by field-space decompression block), arranged as in the schematic architecture (see Fig. \ref{fig:setup}b). In addition, a small stack of Field-Space Attention blocks is placed in the bottleneck: two Field-Space Attention blocks following the encoder and two preceding the decoder. The number of encoder/decoder stages \(N\) is tied to the compression factor via the bottleneck zoom level \(z_{\text{bottleneck}}\): with highest level \(z_{\max}{=}8\), we set \(N = z_{\max}-z_{\text{bottleneck}}\), so that the compression ratio is \(4^{N}\) (e.g., \(z_{\text{bottleneck}}{=}6\) yields \(N{=}2\) and compression factor \(4^{2}{=}16\); \(N{=}3\) corresponds to a factor \(4^{3}{=}64\)). In the final layer of the decoder, we apply the scale conservation operation introduced by Witte et al.\cite{witte2025fieldspaceattentionstructurepreservingearth} to re-enforce the scale-conservative hierarchy before reconstructing the finest field. All Field-Space Autoencoder blocks use the spherical harmonic position embeddings defined in Witte et al.\cite{witte2025fieldspaceattentionstructurepreservingearth}, configured at HEALPix level \(z{=}3\). The models are trained with an RMSE reconstruction loss. The model's hyperparameters are: model dimension \(d_{\text{model}}{=}512\), head dimension \(d_{\text{head}}{=}16\), learning rate \(10^{-3}\), Adam optimizer with 2000 steps of linear warm-up followed by a cosine annealing schedule, a maximum of 30\,000 iterations for single-variable pretraining and 20\,000 iterations for variable-attention fine-tuning, and batch size \(8\).

\bmhead{CNN-VAE Models}
The CNN–VAE family processes fields on the native regular latitude–longitude grid. Its encoder and decoder follow the ResNet-style architecture of Rombach et al., with residual blocks that consist of two convolutional layers with normalization and nonlinearity, wrapped in a skip connection. In the encoder, each spatial downsampling stage comprises two identity ResNet blocks (preserving spatial resolution) followed by a downsampling ResNet block that applies average pooling, reducing the spatial resolution by a factor of 2 in each dimension (overall factor 4). Symmetrically, in the decoder each upsampling stage consists of three identity ResNet blocks followed by an upsampling ResNet block using nearest-neighbour interpolation, increasing the spatial resolution by a factor of 2 in each dimension. Each CNN–VAE implements \(N{+}1\) such downsampling stages in the encoder and \(N{+}1\) upsampling stages in the decoder; in the bottleneck, a global spatial self-attention block (operating on the flattened spatial dimensions) is applied once after the encoder and once before the decoder. The effective compression depth is parameterized by \(N\), with compression ratio \(f = 4^{N}\). For single-variable models, the latent bottleneck has 4 channels; for multi-variable models, the input variables are stacked along the channel dimension and the bottleneck channel count is set to \(4\times\) (number of variables). Because the single- and multi-variable CNN–VAEs require different input and latent channel dimensions, they are implemented as separate models, which is a clear disadvantage compared to our proposed model. CNN–VAEs are trained with an RMSE reconstruction loss plus a Kullback-Leibler-divergence term for latent regularization, as in Rombach et al. \cite{rombach2022high}, with a loss weight of $10^{-10}$. The base channel dimension is \(64\) and doubles with each downsampling stage (mirrored in the decoder), the learning rate is \(5\times 10^{-5}\) (chosen for stability), and we use AdamW with 5000 linear warm-up iterations followed by cosine annealing, a maximum of 60\,000 training iterations, batch size \(4\) for single-variable models, and batch size \(2\) for multi-variable models due to memory constraints.

\bmhead{Compressed Field Diffusion Model}
To enable generative emulation, we train a latent diffusion model that operates directly on the compressed field representations. The model input consists of the learned compressed field state at level $z=5$ and the base average at level $z=3$, on which we define the diffusion and denoising process separately. The backbone architecture utilizes a model dimension of 256 and is composed of four stacked Field-Space Attention blocks designed to process the spatio-temporal structure of the data. Within each block, the model applies variable, spatial, and temporal attention simultaneously, allowing for the concurrent integration of information across physical variables and the global HEALPix grid. Crucially, the temporal attention mechanism operates over an 8-day context window, explicitly accounting for neighboring time steps to enforce temporal consistency in the generated sequence. To condition the generation, the network utilizes three embedding modules: a grid embedder (identical to the Field-Space Autoencoder component) to encode spatial topology, a time embedder to encode calendar timestamps following the implementation by Sun et al.\cite{Sun2023Explicit}, and a diffusion step embedder to signal the current noise level. The diffusion process is trained using a cosine noise scheduler over $T=1000$ steps, with the network objective set to velocity prediction ($v$-prediction)\cite{Zheng2023Improved}. We optimize the model using Adam with a batch size of 16 and a base learning rate of $10^{-3}$. The training schedule consists of 5,000 linear warmup iterations followed by cosine annealing for the remaining 100,000 total iterations. For efficient inference, we employ a Denoising Diffusion Implicit Model (DDIM) sampler\cite{Song2020Denoising} and reduce the sampling trajectory to 100 steps.

\newpage
\begin{appendices}

\section{Multi-variable reconstruction evaluation}

\begin{figure}[!ht]
\def\svgwidth{\linewidth}
\centering
\footnotesize{
\begingroup%
  \makeatletter%
  \providecommand\color[2][]{%
    \errmessage{(Inkscape) Color is used for the text in Inkscape, but the package 'color.sty' is not loaded}%
    \renewcommand\color[2][]{}%
  }%
  \providecommand\transparent[1]{%
    \errmessage{(Inkscape) Transparency is used (non-zero) for the text in Inkscape, but the package 'transparent.sty' is not loaded}%
    \renewcommand\transparent[1]{}%
  }%
  \providecommand\rotatebox[2]{#2}%
  \newcommand*\fsize{\dimexpr\f@size pt\relax}%
  \newcommand*\lineheight[1]{\fontsize{\fsize}{#1\fsize}\selectfont}%
  \ifx\svgwidth\undefined%
    \setlength{\unitlength}{116.35414664bp}%
    \ifx\svgscale\undefined%
      \relax%
    \else%
      \setlength{\unitlength}{\unitlength * \real{\svgscale}}%
    \fi%
  \else%
    \setlength{\unitlength}{\svgwidth}%
  \fi%
  \global\let\svgwidth\undefined%
  \global\let\svgscale\undefined%
  \makeatother%
  \begin{picture}(1,1.2128597)%
    \lineheight{1}%
    \setlength\tabcolsep{0pt}%
    \put(0.69996332,1.18765517){\color[rgb]{0,0,0}\makebox(0,0)[t]{\lineheight{1.25}\smash{\begin{tabular}[t]{c}{\fontsize{8pt}{8pt} CNN-VAE 64x}\end{tabular}}}}%
    \put(0.2378135,1.18765519){\color[rgb]{0,0,0}\makebox(0,0)[t]{\lineheight{1.25}\smash{\begin{tabular}[t]{c}{\fontsize{8pt}{8pt} FS-AE 64x}\end{tabular}}}}%
    \put(0,0){\includegraphics[width=\unitlength,page=1]{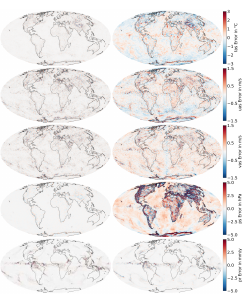}}%
  \end{picture}%
\endgroup%
}
\caption{\textbf{Multi-variable reconstruction Error.} Absolute Errors of all five variables at 64× compression from a selected sample, comparing Errors from the Field-Space Autoencoder (FS-AE) and the CNN-VAE baseline to ground truth. Positive values correspond to overestimation, negative values to underestimation.}
\label{fig:allvar_plots}
\end{figure}

\newpage

\section{RMSE comparison for zero-shot super-resolution.}
\begin{figure}[!ht]
\def\svgwidth{\linewidth}
\centering
\includegraphics[width=0.6\textwidth]{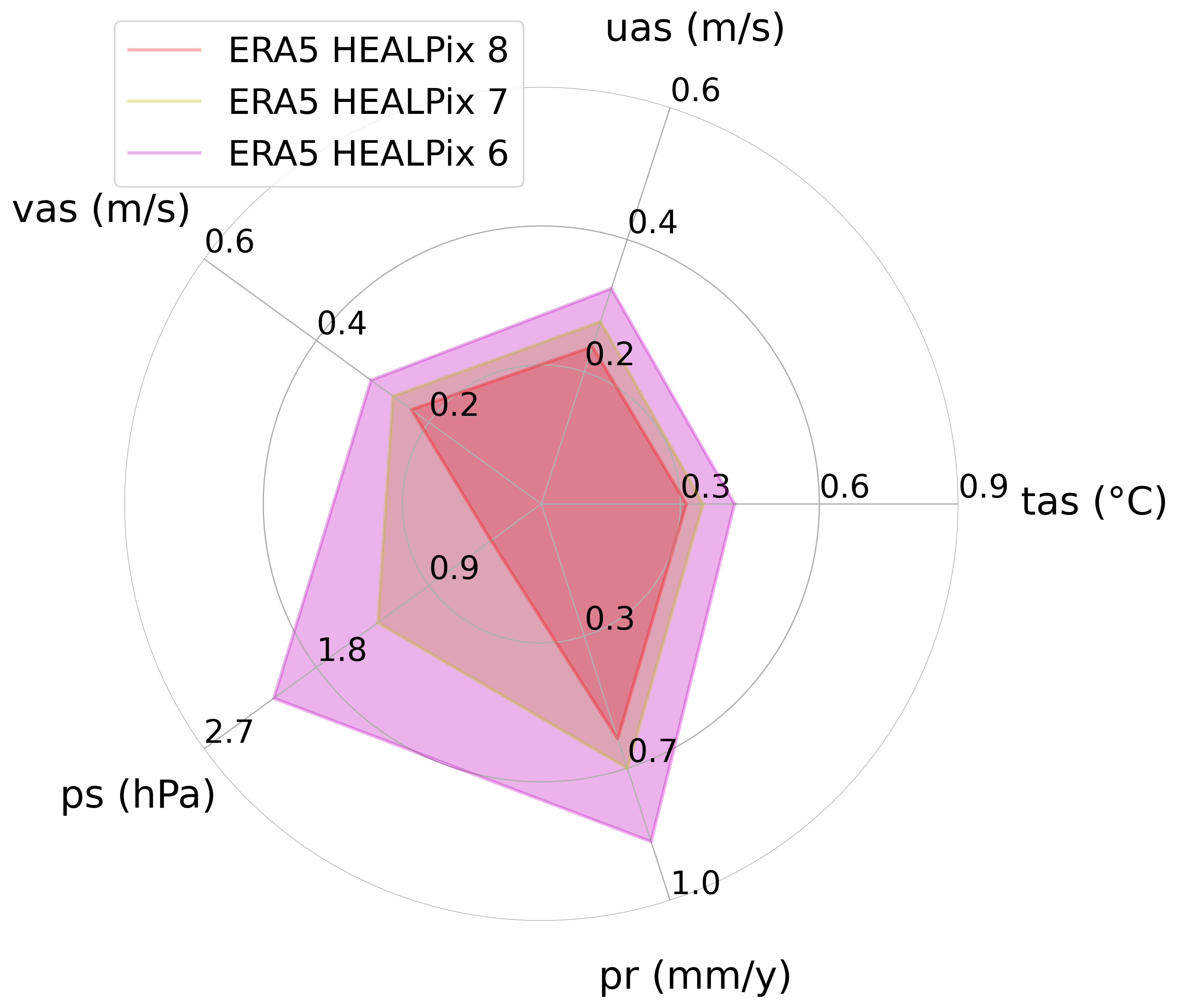}
\caption{\textbf{RMSE comparison for zero-shot super-resolution.} Radar chart displaying root-mean-square error (RMSE) for five variables ($tas$, $uas$, $vas$, $ps$, $pr$) using the Field-Space Autoencoder ($f=64$) model. Colored polygons represent three decoding settings: native resolution compression (red) at HEALPix level 8, zero-shot super-resolution at HEALPix level 7 (yellow), and zero-shot super-resolution at HEALPix level 6 (magenta). Radial axes have independent scales corresponding to the magnitude of each variable.
}
\label{fig:radar_sr}
\end{figure}

\newpage

\section{Multi-Variable Reconstructions from MPI-ESM model input.}
\begin{figure}[!ht]
\def\svgwidth{0.95\linewidth}
\centering
\footnotesize{
\begingroup%
  \makeatletter%
  \providecommand\color[2][]{%
    \errmessage{(Inkscape) Color is used for the text in Inkscape, but the package 'color.sty' is not loaded}%
    \renewcommand\color[2][]{}%
  }%
  \providecommand\transparent[1]{%
    \errmessage{(Inkscape) Transparency is used (non-zero) for the text in Inkscape, but the package 'transparent.sty' is not loaded}%
    \renewcommand\transparent[1]{}%
  }%
  \providecommand\rotatebox[2]{#2}%
  \newcommand*\fsize{\dimexpr\f@size pt\relax}%
  \newcommand*\lineheight[1]{\fontsize{\fsize}{#1\fsize}\selectfont}%
  \ifx\svgwidth\undefined%
    \setlength{\unitlength}{233.74189482bp}%
    \ifx\svgscale\undefined%
      \relax%
    \else%
      \setlength{\unitlength}{\unitlength * \real{\svgscale}}%
    \fi%
  \else%
    \setlength{\unitlength}{\svgwidth}%
  \fi%
  \global\let\svgwidth\undefined%
  \global\let\svgscale\undefined%
  \makeatother%
  \begin{picture}(1,1.19086267)%
    \lineheight{1}%
    \setlength\tabcolsep{0pt}%
    \put(0.22650535,1.17807527){\color[rgb]{0,0,0}\makebox(0,0)[t]{\lineheight{1.25}\smash{\begin{tabular}[t]{c}\scriptsize{MPI-ESM}\end{tabular}}}}%
    \put(0.6872346,1.17782509){\color[rgb]{0,0,0}\makebox(0,0)[t]{\lineheight{1.25}\smash{\begin{tabular}[t]{c}\scriptsize{FS-AE ($f=64$)}\end{tabular}}}}%
    \put(0,0){\includegraphics[width=\unitlength,page=1]{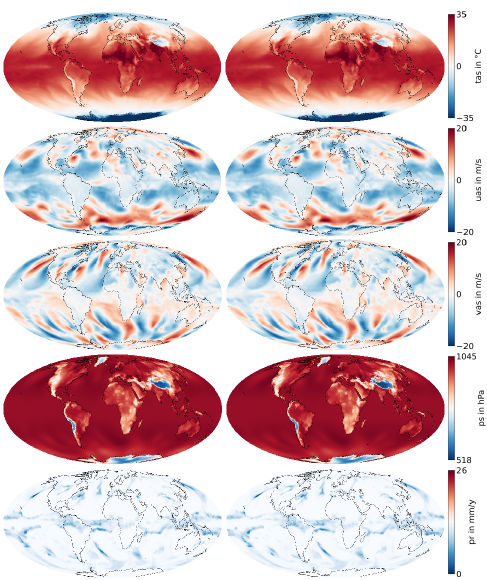}}%
  \end{picture}%
\endgroup%
}
\caption{\textbf{Multi-variable reconstructions from the MPI-ESM Model.} Example global reconstructions of all five variables, comparing outputs from the Field-Space Autoencoder (FS-AE) with compression ratio $f=64$ and the MPI-ESM ground truth, which were given as input.
}
\label{fig:eval_sr_allvar}
\end{figure}

\newpage

\section{Angular power spectra of HEALPix surface temperature fields for super-resolution}
\begin{figure}[!ht]
\def\svgwidth{\linewidth}
\centering
\includegraphics[width=1.0\textwidth]{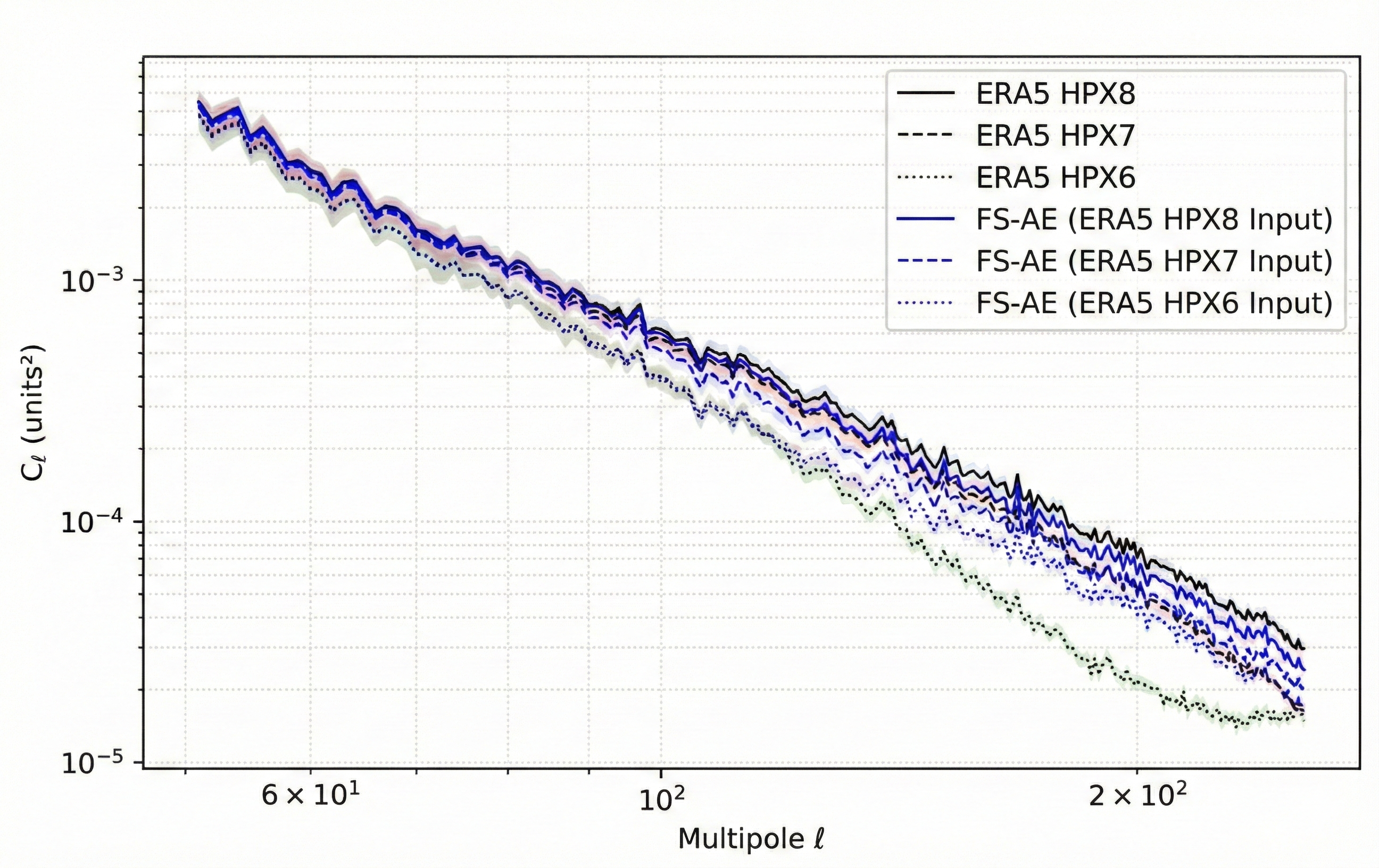}
\caption{\textbf{Comparing the angular power spectra of surface temperature fields for super-resolution.} For each dataset,  $C_\ell$ is computed per timestep from the surface temperature maps and then averaged across time; the mean spectrum is shown as a solid curve on log–log axes. The shaded areas indicate the temporal variability ($±1\sigma$ across timesteps). Higher multipoles $\ell$ represent finer spatial scales; lower $\ell$ capture large-scale temperature structure. Units of $C_\ell$ reflect the native temperature units of the input fields (°C).
}
\label{fig:spetral_analysis_sr}
\end{figure}

\newpage

\section{Comparing the angular power spectra of surface temperature fields from MPI-ESM}
\begin{figure}[!ht]
\def\svgwidth{\linewidth}
\centering
\includegraphics[width=1.0\textwidth]{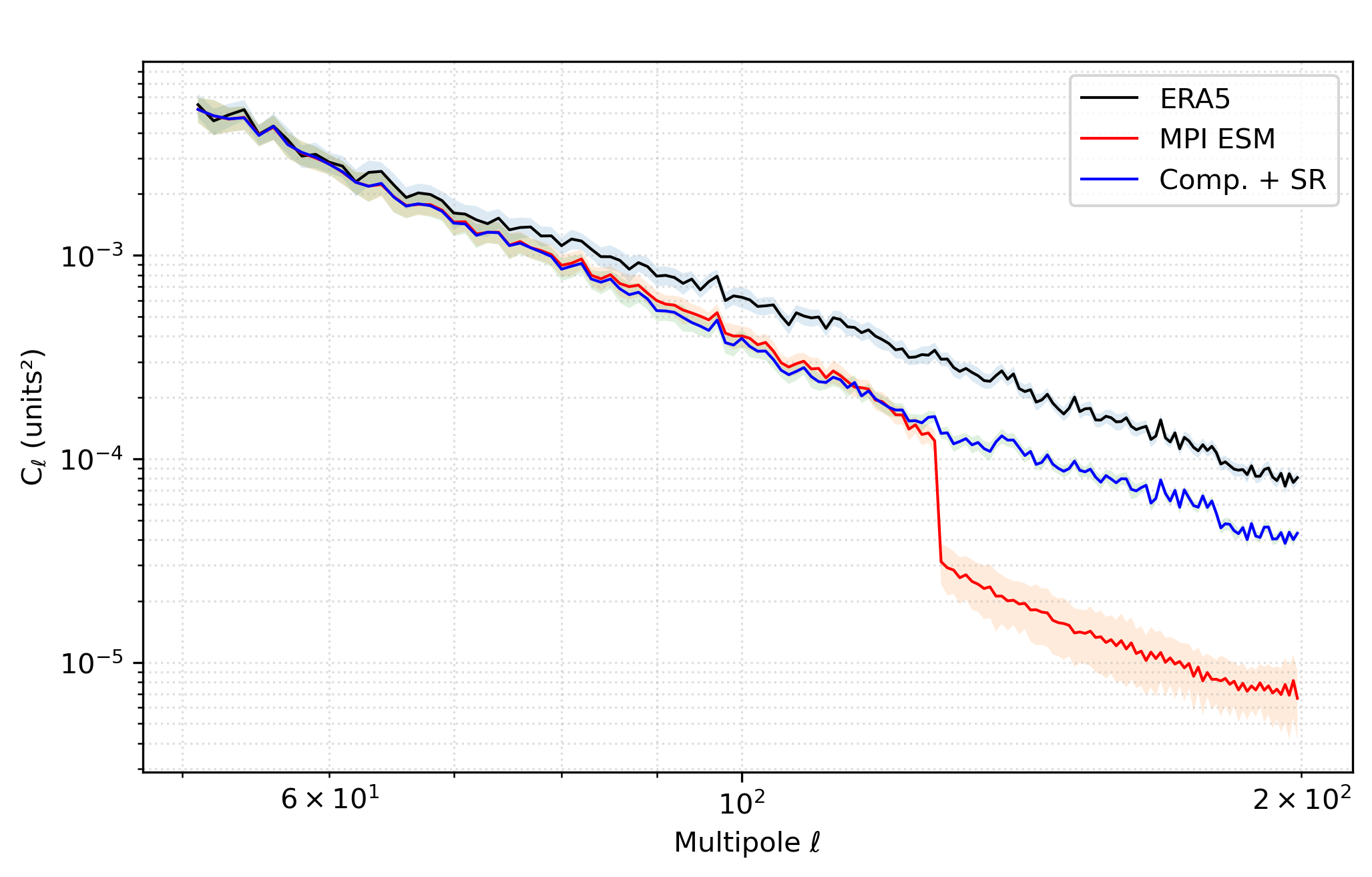}
\caption{\textbf{Angular power spectra of surface temperature fields over time.} For each dataset (ERA5, MPI ESM, Comp. + SR),  $C_\ell$ is computed per timestep from the surface temperature maps and then averaged across time; the mean spectrum is shown as a solid curve on log–log axes. The shaded areas indicate the temporal variability ($±1\sigma$ across timesteps). Higher multipoles $\ell$ represent finer spatial scales; lower $\ell$ capture large-scale temperature structure. Units of $C_\ell$ reflect the native temperature units of the input fields (°C).
}
\label{fig:spetral_analysis_temp_mpiesm}
\end{figure}

\newpage

\section{Comparing the angular power spectra of surface u-wind fields from MPI-ESM}
\begin{figure}[!ht]
\def\svgwidth{\linewidth}
\centering
\includegraphics[width=1.0\textwidth]{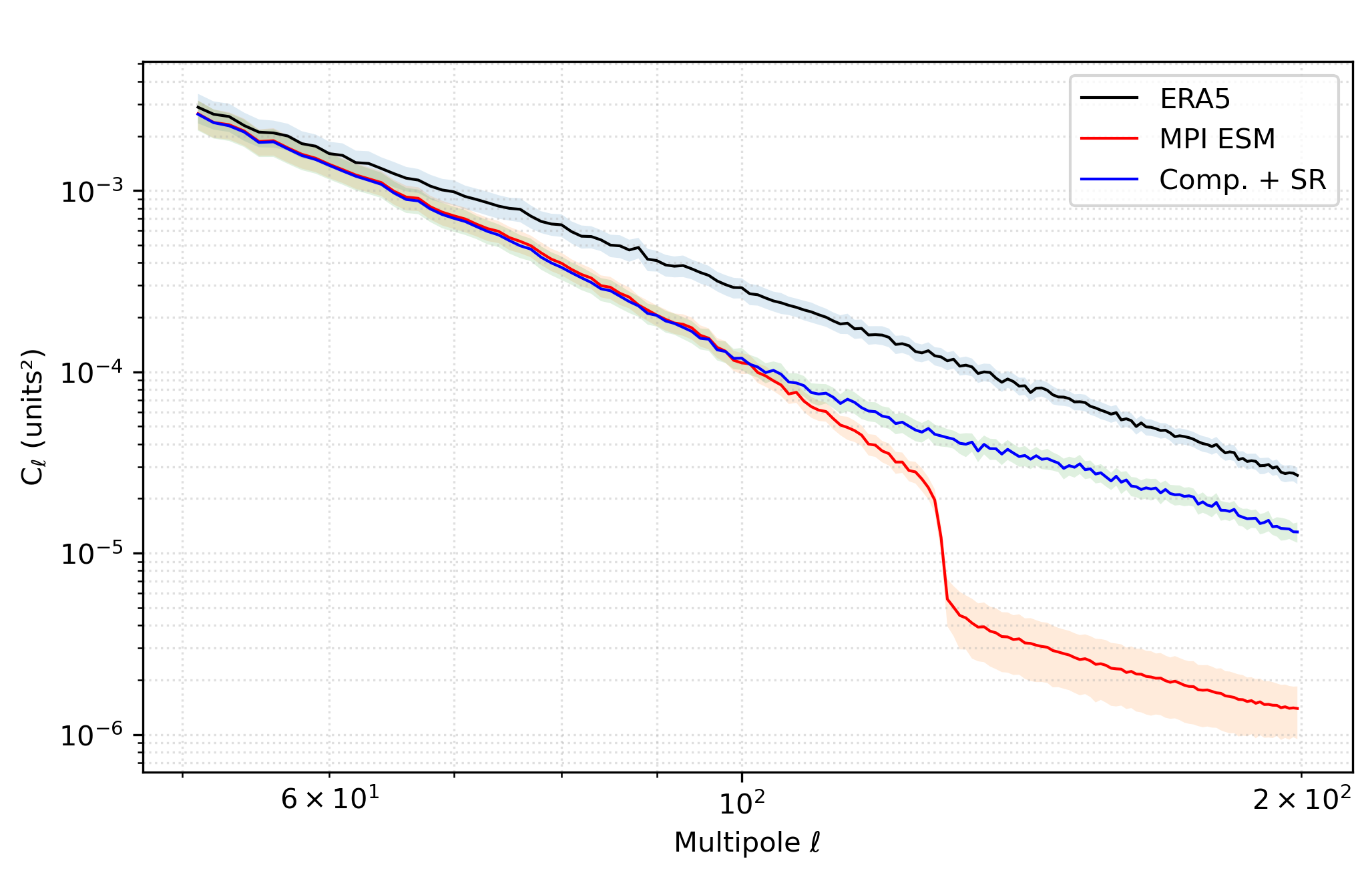}
\caption{\textbf{Angular power spectra of surface temperature fields over time.} For each dataset (ERA5, MPI ESM, Comp. + SR),  $C_\ell$ is computed per timestep from the surface u-wind maps and then averaged across time; the mean spectrum is shown as a solid curve on log–log axes. The shaded areas indicate the temporal variability ($±1\sigma$ across timesteps). Higher multipoles $\ell$ represent finer spatial scales; lower $\ell$ capture large-scale u-wind structure. Units of $C_\ell$ reflect the native u-wind units of the input fields (m/s).
}
\label{fig:spetral_analysis_uas_mpiesm}
\end{figure}

\newpage




\end{appendices}


\bibliography{sn-bibliography}

\end{document}